\documentclass[10pt,twocolumn,letterpaper]{article}
\usepackage[pagenumbers]{cvpr}
\usepackage[table]{xcolor}
\usepackage[accsupp]{axessibility}

\usepackage{booktabs}  
\usepackage{graphicx, amsmath, amssymb, multirow, overpic, textpos}
\usepackage{array, color}
\usepackage{amsmath}
\usepackage{bbm}

\usepackage{pifont}

\newcommand{\xmark}{\text{\ding{55}}}

\definecolor{citecolor}{HTML}{0071bc}
\definecolor{color_ao}{gray}{0.5}
\definecolor{color_our}{rgb}{0.66,0.82,0.56}
\definecolor{color_pre}{rgb}{0.52,0.59,0.69}
\definecolor{Gray}{gray}{0.9}
\definecolor{LighterGray}{gray}{0.93}
\definecolor{LightGrayForTableRule}{gray}{0.92}
\definecolor{DarkGray}{gray}{0.5}
\definecolor{Black}{rgb}{0.0, 0.0, 0.0}
\definecolor{NiceBlue}{rgb}{0.11764705882352941, 0.5647058823529412, 1.0}
\definecolor{NiceGreen}{rgb}{0.0, 0.5, 0.0}

\definecolor{citecolor}{HTML}{0071BC}
\definecolor{linkcolor}{HTML}{ED1C24}
\usepackage{tabularx}

\definecolor{LightGrayForTableRule}{gray}{0.92}

\newcommand{\authorskip}{\hspace{5mm}}
\usepackage[hang,flushmargin]{footmisc} 

\makeatletter
\newcommand{\footnotenosuper}[1]{%
   \def\@makefnmark{\hbox{}}
   \footnote{#1}
   \def\@makefnmark{\hbox{\@textsuperscript{\normalfont\@thefnmark}}}
}
\makeatother

\usepackage{marvosym}

\usepackage{color}
\usepackage{cinzel}

\definecolor{verylightblue}{RGB}{220,235,245}
\newcommand{\baseline}[1]{\cellcolor{verylightblue}{#1}}

\usepackage[most]{tcolorbox} 
\usepackage{xcolor} 
\usepackage{lipsum}

\newtcolorbox{UserBox}[2][]{
    colback=green!5!white,
    colframe=green!55!black,
    coltext=black,
    boxsep=1pt,
    arc=5pt,
    auto outer arc,
    left=5pt,
    right=5pt,
    top=1pt,
    bottom=1pt,
    box align=top,
    title=#2,
    #1
}
\newtcolorbox{AssistantBox}[2][]{
    colback=orange!5!white,
    colframe=orange!75!black,
    coltext=blue!25!black,
    boxsep=1pt,
    arc=5pt,
    auto outer arc,
    left=5pt,
    right=5pt,
    top=1pt,
    bottom=1pt,
    box align=top,
    title=#2,
    #1
}
\newtcolorbox{VideoInputBox}{
    colback=gray!5!white,
    colframe=gray!75!black,
    coltext=gray!25!black,
    boxsep=5pt,
    arc=5pt,
    left=5pt,
    right=5pt,
    top=0.5pt,
    bottom=0.5pt,
    box align=top,
    before upper=\hspace{-1em}\textbf{Streaming video input: }
}

\definecolor{cvprblue}{rgb}{0.21,0.49,0.74}
\usepackage[pagebackref,breaklinks,colorlinks,citecolor=cvprblue]{hyperref}
\usepackage[accsupp]{axessibility}

\def\paperID{3670} 
\def\confName{CVPR}
\def\confYear{2024}

\title{VideoLLM-online: Online Video Large Language Model for Streaming Video}
\author{Joya Chen$^1$ \authorskip Zhaoyang Lv$^2$ \authorskip Shiwei Wu$^{1}$  \authorskip Kevin Qinghong Lin$^1$ 
\authorskip Chenan Song$^1$ \\ \authorskip Difei Gao$^1$ \authorskip Jia-Wei Liu$^1$ \authorskip Ziteng Gao$^1$ \authorskip Dongxing Mao$^1$ \authorskip Mike Zheng Shou$^{1}$\textsuperscript{\Letter}\\[1mm]
$^1$Show Lab, National University of Singapore \quad $^2$Reality Labs Research, Meta
}
\begin{document}
\twocolumn[{
\maketitle
\vspace{-3em}
\renewcommand\twocolumn[1][]{#1}
\begin{center}
    \centering
    \includegraphics[width=1.0\textwidth]{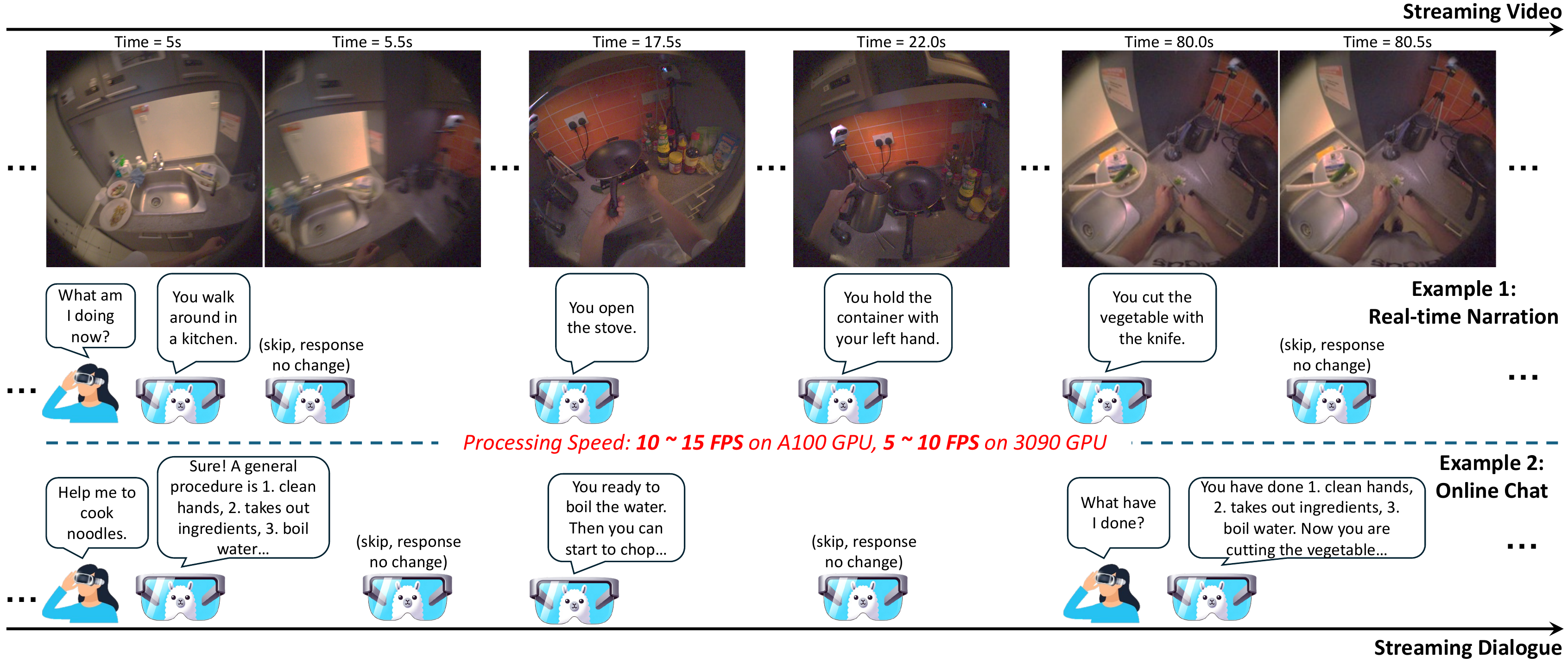}
    \captionof{figure}{
    Zero-shot examples of our VideoLLM-online applied to an egocentric video stream from Ego-Exo4D dataset~\cite{egoexo4d}. Our model is designed for temporally aligned, long-context, real-time dialogue in continuous video streams, shedding light on the future always-on, contextual AI assistants (\eg, smart AR glasses). Model responses are appropriately simplified for better visualization.}
    \label{figure:teaser}
\end{center}
}]

\quad \footnotenosuper{\textsuperscript{\Letter}Corresponding Author.\\\textit{This is the Llama-3 upgraded version for CVPR camera-ready.}}

\vspace{-3mm}
\begin{abstract}
Recent Large Language Models have been enhanced with vision capabilities, enabling them to comprehend images, videos, and interleaved vision-language content. However, the learning methods of these large multimodal models typically treat videos as predetermined clips, making them less effective and efficient at handling streaming video inputs. In this paper, we propose a novel Learning-In-Video-Stream (LIVE) framework, which enables temporally aligned, long-context, and real-time conversation within a continuous video stream. Our LIVE framework comprises comprehensive approaches to achieve video streaming dialogue, encompassing: (1) a training objective designed to perform language modeling for continuous streaming inputs, (2) a data generation scheme that converts offline temporal annotations into a streaming dialogue format, and (3) an optimized inference pipeline to speed up the model responses in real-world video streams. With our LIVE framework, we built VideoLLM-online model upon Llama-2/Llama-3 and demonstrate its significant advantages in processing streaming videos. For instance, on average, our model can support streaming dialogue in a 5-minute video clip at over 10 FPS on an A100 GPU. Moreover, it also showcases state-of-the-art performance on public offline video benchmarks, such as recognition, captioning, and forecasting. The code, model, data, and demo have been made available at \href{https://showlab.github.io/videollm-online}{showlab.github.io/videollm-online}.
\end{abstract}
    
\begin{figure*}[t]
    \centering
    \includegraphics[width=\linewidth]{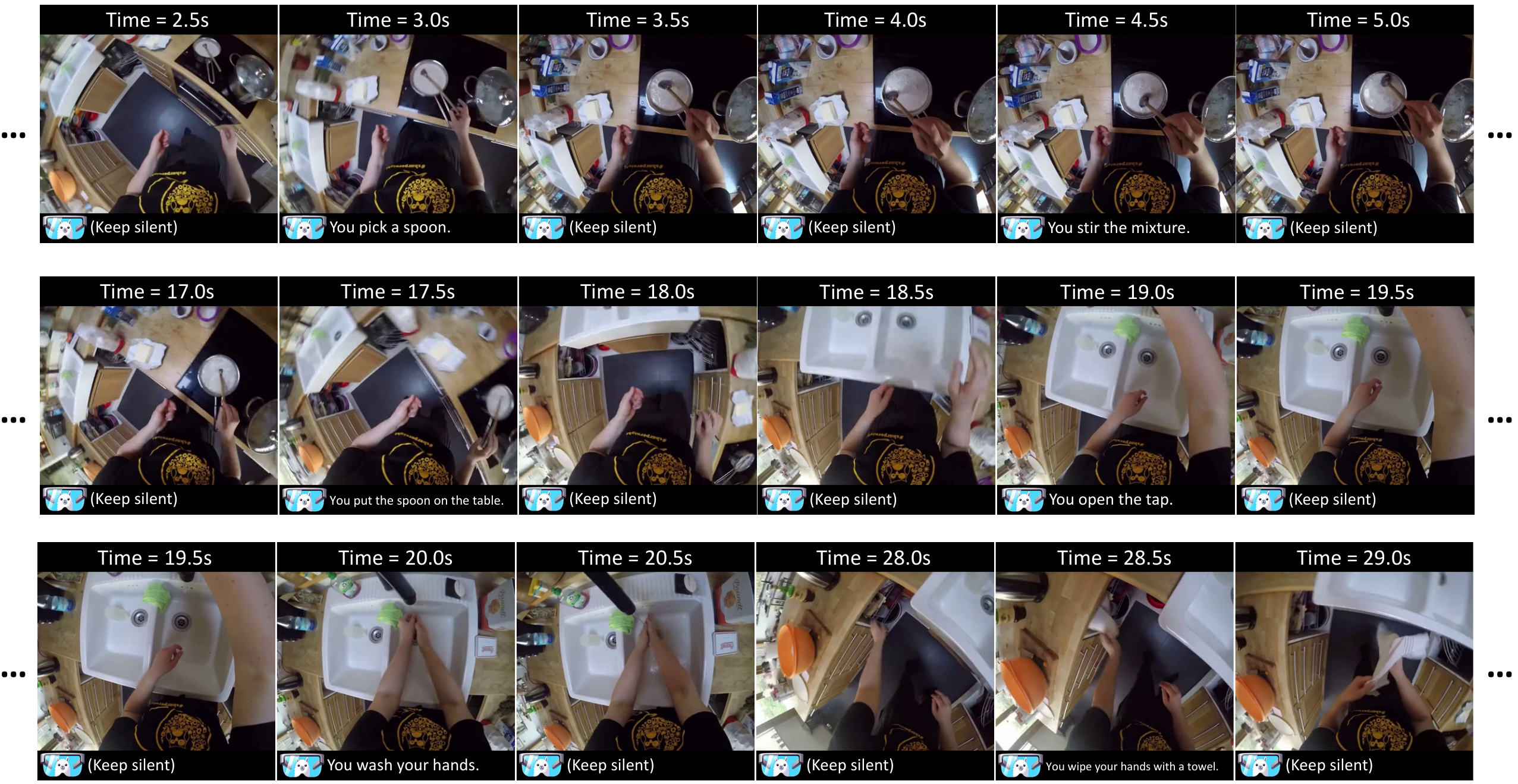}
    \caption{Our model shows strong temporal alignment capability in streaming video narration. The query at the beginning is ``Please describe what I am doing in real time".}
    \label{figure:align}
\end{figure*}

\section{Introduction}
\label{section:introdcution}

Building the future of an always-on, contextual AI assistant that can actively help humans in various situations, digitize inputs as episodic memories, and forecast future plans in an online, continuous setting represents one of the "holy grail" missions in AI research.
Powered by advancements in large language models (LLMs)~\cite{scaling_law,gpt3,instructgpt,chatgpt,llama1,gpt4}, recent large multimodal models (LMMs) have unveiled impressive capabilities such as vision-language dialogue~\cite{blip2,instructblip,qwen_vl,llava,llava1.5,minigpt4,otter}, spatial understanding~\cite{gpt4roi,ferret,qwen_vl,macaw_llm,lisa,kosmos-2}, processing diverse modalities~\cite{dreamllm,nextgpt,chameleon_meta,anymal,imagebind_llm}. Seminal exemplars, like OpenAI's GPT-4V~\cite{gpt4v} or GPT-4o~\cite{gpt4o}, are progressively evolving into highly versatile human AI assistants. 

However, even the most advanced GPT-4o~\cite{gpt4o} has only achieved streaming voice-driven multimodal assistance.\footnote{Based on their demo videos, the GPT-4o responses to visual scenes can only occur after an active human voice input.} Therefore, it is time to envision an always-on, contextual, J.A.R.V.I.S-like video assistant that supports free-form user-assistant dialogue within the video stream, which we term ``video streaming dialogue''. Unlike existing LMMs for video understanding (\ie, VideoLLMs)~\cite{video_chatgpt,videochat,vid2seq,videollama,moviechat,embodiedgpt,videollava} that work offline with manually selected short-video clips, an online assistant should continuously receive video frames with visual content that is constantly refreshed. This paradigm shift presents new challenges. 
First, the user query may come with \textbf{temporally aligned} requirements (\eg, ``alert me when it's time to flip the steak''), thus the VideoLLM should scan every incoming frame to avoid event missing, instead of only yielding video-level responses.
Second, to answer questions regarding summarization and planning, the VideoLLM must retain the \textbf{long-context} historical vision and language, which poses huge risk of exceeding maximum context window of LLMs, as well as introduces considerable burden to causal decoding speed and GPU memory.
Third, the VideoLLM should generate the answer in \textbf{real-time}, keeping pace with the video stream for always-on scenario. These abilities, however, are even partially overlooked by the most advanced AI assistants~\cite{gpt4v,gemini}.

One possible path towards such an online VideoLLM, inspired by current interleaved vision-language models~\cite{deepspeed_visualchat,gpt4v,gemini,qwen_vl,otter}, is to employ a multi-turn dialogue format to achieve per-frame chatting within a video stream. This can be accomplished by facilitating very frequent user interactions, utilizing the visual frame as query at each timestamp to obtain the answer. We follow this to perform prompt engineering for GPT-4V~\cite{gpt4v,mmvid}, but the results are disappointing: GPT-4V tends to output lengthy content at every frame, leading to significant delays, making it impractical for real-time streaming video. We also explore training baseline models for per-frame chatting. Unfortunately, this approach evidently diminishes the language modeling capability, likely due to harmful language modeling on an excessive number of redundant frames.

We propose Learning-In-Video-strEam (LIVE), a comprehensive framework that encompasses learning, data, and inference methods to develop an online video assistant. Unlike per-frame dialogue approach, LIVE introduces a novel training objective termed \textit{Streaming EOS (End-Of-Sequence) prediction} that enables the model to learn when to response or remain silent in a video stream. This objective differs from next-token prediction since EOS tokens here will not appear in the input/output sequence. However, it can work well with the autoregressive loss to train an online VideoLLM. This design reduces unnecessary context, helping the model to manage much longer streaming videos. Nevertheless, the training still requires data from user queries and assistant responses within video streams, which is scarce in popular video datasets usually used for training offline video models. To address this issue, LIVE presents a streaming dialogue generation scheme that converts offline annotations into online dialogues to support free-form chatting. To enhance inference efficiency, LIVE leverages continuous key-value caching for streaming assistance, and parallelizes the fast visual encoding and slow language decoding to prevent bottlenecks, thus moving towards real-time application.

With LIVE framework, we build a simple VideoLLM-online model upon CLIP~\cite{clip} vision encoder and Llama-2~\cite{llama2}/Llama-3~\cite{llama3} language model. To evaluate the performance of video streaming dialogue, we utilize the language perplexity metric and design two new metrics to comprehensively assess the model's capabilities in language modeling, temporal responsiveness, and overall streaming fluency. Experiments with real-time Ego4D narration~\cite{ego4d} demonstrate that our method shows advantages in all that metrics, with higher speed and lower memory cost. Furthermore, our model achieves state-of-the-art results on numerous offline benchmarks, such as short- and long-term activity recognition and forecasting on the COIN and Ego4D LTA benchmarks~\cite{coin,ego4d}. In addition, our model has good speed/memory efficiency, \eg, allowing continuous 5-minute video streaming dialogue with memory cost less than 20 GB and average speed higher than 10 FPS on a single A100 GPU, paving the way for future real-world usage.




 
\section{Related Work}

\noindent\textbf{Visual Dialogue.} Before transformers~\cite{transformer} become mainstream in vision, visual dialogue methods~\cite{visdial,viddial} tend to employ a visually enhanced encoder, with a discriminative head to select candidate answers or a recurrent architecture to generate multi-turn responses. For the encoder, a variety of attention mechanism-based approaches~\cite{dan,fga,ltmi} have been proposed to learn the interactions between the image, the answers, and the dialogue history. There have also been explorations of encoder-only BERT~\cite{bert} models for visual dialogue~\cite{vdbert,large_visual_dialogue}. However, most of them rely solely on a single image/video at the beginning of the conversation, followed by multi-turn pure language dialogue, which makes them less flexible than the current interleaved vision-language dialogue systems.

\noindent\textbf{Large Multimodal Models.} 
The advent of large language models (LLMs)~\cite{gpt3,instructgpt,chatgpt} has inspired a series of LLM multimodal variants, \ie large multimodal models (LMMs).
Early LMMs~\cite{flamingo,blip2,llava,minigpt4,instructblip} achieve image dialogue by projecting the image encoding (\eg, from CLIP~\cite{clip}) to align with LLM embedding space. Then, lots of efforts~\cite{gpt4v, qwen_vl, otter, gemini,deepspeed_visualchat} explore more free-form interleaved vision-text chatting, spatial understanding~\cite{shikra, gpt4roi,ferret, minigpt4v2, qwen_vl,macaw_llm,lisa}, video comprehension~\cite{video_chatgpt,videochat,vid2seq,videollama,moviechat,embodiedgpt,videollava}, etc.
However, when it comes to online scenario, there is less exploration on how LLMs can fulfill the temporal alignment, long-context, and real-time requirements for streaming video inputs. Our research bridges this gap, offering comprehensive solutions across model training, data, and inference to study the problem.
 
\noindent\textbf{Online Video Understanding.} 
Typical video understanding benchmarks, such as action recognition~\cite{i3d}, temporal action localization~\cite{activitynet}, video question answering~\cite{tvqa}, and video dialogue~\cite{viddial}, typically allow models to access entire video frames to make predictions, a setting referred to as "offline". However, such setting does not align well with many real-time demands (\eg, autonomous driving, AR glasses). Instead, there is a growing focus on ``online'' video understanding problems like online action detection~\cite{oad} and anticipation~\cite{anticipation}, which aim to identify the current/future action at each timestamp without seeing the future. Our study pioneers LMMs for online video understanding. Unlike previous online action detection~\cite{testra,oadtr} or anticipation models~\cite{avt,antgpt}, which primarily address one task with a highly customized model, we aim to propose a general solution to achieve free-form dialogue along the online video stream, enabling a model to flexibly handle diverse tasks. Streaming video caption~\cite{stream_dvc} belongs to our concurrent work, but it only supports captioning rather than free-form dialogue, and its streaming caption temporal region is fixed, making it much less flexible and general than our work.

\noindent\textbf{Efficient Token Decoding.} 
Efficient token decoding for LLMs and LMMs is essential for applying them to real-time online services. To accelerate that, a diversity of strategies has been proposed, such as parallelism on batch dimension~\cite{flexgen} or cached key-value sequence~\cite{flash_decoding}, computation/memory management optimization~\cite{paged_attention,flexgen,fastgen,orca,flash_decoding++}, and even some lossy approaches~\cite{streamingllm,speculative_decoding}. Our focused streaming video scenario has less concerns in large batch processing, but expects faster decoding to avoid excessive frame skipping. We have considered this in training objectives, and further propose some inference schemes to accelerate the decoding efficiency.
\section{Method}
In this section, we present Learning-In-Video-strEam (LIVE) framework that enables LMMs to provide temporally aligned response, handle long-context streaming video, and run efficiently towards real-time usage. We will start from the problem definition of ``video streaming dialogue'', analyze its challenges, and introduce our approach to solve the problem.

\subsection{Video Streaming Dialogue}\label{section:video_streaming_dialogue}

\noindent\textbf{Problem Formulation.}
Though huge successes have been witnessed in large multimodal models (LMMs), the assistance scenario like a smart AR glass helping the user with cooking, is still far from the capabilities of current LMMs, even for the most advanced version, GPT-4V~\cite{gpt4v}. For example, despite carefully prompting GPT-4V similarly to MMVid~\cite{mmvid}---to perform per-frame dialogue for handling streaming video inputs---the redundancy in responses between frames, a limited context window of 10$\sim$50 frames, and slow speed, collectively render the current GPT-4V unsuitable for online video understanding, as our prompting analysis demonstrates (see supplementary material).














To bridge the gap, we define a problem termed ``video streaming dialogue''. Given the context sequence before time $t=t_1$, denoted as $\texttt{[Ctx$^{t<t_1}$]}$, which may encompass previous vision-language content (\eg, historical user queries, video frames, assistant responses), and an ongoing continuous video stream from $t_1$ to $t_2$, denoted as $\texttt{[Frame$^{t_1\leq t\leq t_2}$]}$, our goal is (1) to determine whether the current time $t_2$ is suitable for language modeling; (2) to carry out language modeling 

\begin{equation}
    \max P(\texttt{[Txt$^{t_2}_{i+1}$]}|\texttt{[Ctx$^{<t_1}$]}, \texttt{[F$^{t_1\leq t\leq t_2}$]}, \texttt{[Txt$^{t_2}_{\leq i}$]})
\end{equation}

\noindent if $t_2$ is determined, where \texttt{[Txt$^{t}_{i}$]} denotes the ideal language token in the $i$-th position at timestamp $t$. \texttt{[F]} is the abbreviation of \texttt{[Frame]}.

In the following, we analyze if existing techniques have been enough to solve the problem.

\noindent\textbf{Interleaved/Per-frame Dialogue are Suboptimal}. First, we investigate whether the popular approach of interleaved vision-language chatting can address this problem. Related to our formulation above, such a method learns language modeling after given frames between timestamps $t_1$ and $t_2$. However, if this approach is adopted during inference, it necessitates the manual selection of timestamps $t_1$ and $t_2$, which does not align with the concept of video streaming dialogue. Current VideoLLMs~\cite{video_chatgpt,videochat,vid2seq,videollama,moviechat,embodiedgpt,videollava} represent a simplified version of this approach, engaging in single- or multi-turn pure language dialogue following video clip inputs.

If we consider a more free-form format, multi-turn interleaved vision-language chatting~\cite{gpt4v,deepspeed_visualchat,qwen_vl,gemini,otter}, we can get a per-frame chatting solution that might be hopeful to solve video streaming dialogue, which performs per-frame language modeling

\begin{equation}
    \max P(\texttt{[Txt$^{t}_{i+1}$]}|\texttt{[Ctx$^{<t_1}$]}, \texttt{[Frame$^{t}$]}, \texttt{[Txt$^{t}_{\leq i}$]})
\end{equation}

\noindent for every frame from timestamp $t_1 \leq t \leq t_2$. Since $t_2$ is necessary to output answer, so we can simply learn short text (\eg, ``this is not the time to answer'') between $t_1 \leq t < t_2$. However, this approach imposes a significant burden on processing speed. For every frame, performing the slow, recurrent, and lengthy next-token prediction with a billion-scale language model makes it extremely hard to achieve real-time video streaming dialogue. Then, this will lead to unavoidable frame skipping, which is unexpected and detrimental for temporal alignment. Furthermore, it taxes the limited context window of the language model, presenting challenges for modeling long contexts and managing GPU memory efficiently.

\noindent\textbf{Streaming EOS Prediction}. To solve the problem mentioned above, we first consider a more efficient per-frame chatting method: simply assigning the End-of-Sequence (EOS) token as the content for chatting between $t_1 \leq t < t_2$. However, this approach remains suboptimal. The dialogue prompt template (\eg, \texttt{[INST]}, \texttt{[/INST]}, space tokens in Llama~\cite{llama1,llama2}) still consumes a considerable number of tokens per frame, which is less favorable due to the numerous frames in streaming video. Furthermore, the excessive number of EOS tokens in the sequence can significantly increase the language model's perplexity, as we have observed in our experiments.

Instead, we propose a novel training objective named ``streaming EOS prediction'' to address this issue. We still assume $t_2$ is essential for decoding language; thus, we normally learn language modeling here:

\begin{equation}
    \max P(\texttt{[Txt$^{t_2}_{i+1}$]}|\texttt{[Ctx$^{<t_2}$]}, \texttt{[Frame$^{t_2}$]}, \texttt{[Txt$^{t_2}_{\leq i}$]}).
\end{equation}

\noindent However, for timestamps $t_1 \leq t < t_2$, which are redundant for producing answers, we directly learn the model to predict EOS token on the frame tokens, \ie

\begin{equation}
    \max P(\text{EOS}|\texttt{[Ctx$^{<t}$]}, \texttt{[Frame$^{t}$]}), \text{where } t_1 \leq t < t_2.
\end{equation}

\noindent In this way, we ``skip'' a dialogue turn and learn to determine when it is appropriate to decode language for streaming inputs. During inference, if EOS is predicted on a frame, then we can directly ask the next frame to input.  Meanwhile, the EOS token is not appended to the context to prevent it from affecting the language modeling. Therefore, this task is not about next-token prediction; however, it can work with autoregressive loss to train a video streaming dialogue model. In the following, we first introduce how can we get the data of streaming video and timestamped language annotations, then present the details about the model and the training procedure.

We also note that the EOS token mentioned here is not limited to the real EOS token used in language models (\eg, \texttt{</s>} in Llama). It is permissible to use any token or to introduce a new token, provided that it is specified in the system prompt. We use this term solely for simplicity.

\subsection{Data}

\noindent\textbf{Online Annotations to Video Streaming Dialogue.} Some video datasets, such as Ego4D narrations~\cite{ego4d}, are inherently collected in a streaming manner, with annotators providing real-time narrations as they watch a 5-minute long video clip. However, prior research~\cite{egovlp,lavila} has predominantly focused on learning from short, discrete clips (\eg only 32 frames), rather than in a continuous streaming context. For this dataset, we followed the same instructions provided to human annotators~\cite{ego4d} as our model training prompt. This prompt instructs the model to simulate human annotators to streamingly generate narrations in a 5-minute video (about 600 frames in 2 FPS). For demo purposes (not for experiments), we also utilize Llama-2-13B-Chat~\cite{llama2} or Llama-3-8B-Instruct~\cite{llama3} to rephrase the narration text, correcting grammatical errors and typos, and converting it into a more understandable version (\eg, changing ``C does...'' to ``You do...''). 

\begin{figure}[!t]
    \centering
    \includegraphics[width=1.0\linewidth]{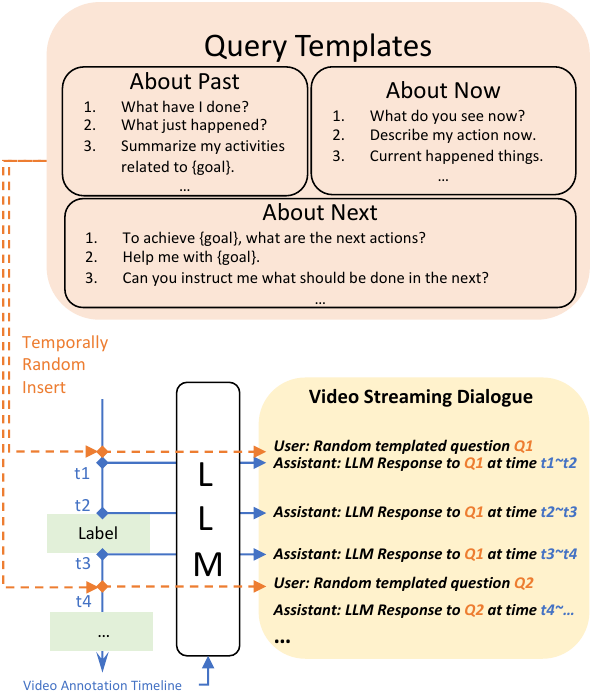}
    \caption{\textbf{The streaming dialogue data generation method in our LIVE framework}. We randomly insert templated questions into the video timeline and ``expose'' the ground-truth video annotations (along with their timestamps) to LLMs, prompting them to answer the queries  within a period of time.} 
    \label{figure:data}
\end{figure}

\noindent\textbf{Offline Annotations to Video Streaming Dialogue.}
Despite the Ego4D narration data being collected in a streaming manner, most prevalent video datasets~\cite{ego4d,howto100m,ek100,coin} are used to train offline models and only feature temporal segment annotations paired with basic language descriptions (\eg, activities, narrations). To bridge this gap, we propose a method for synthesizing dialogue data from these sources. As shown in Figure~\ref{figure:data}, our key idea is use LLM to generate user-assistant dialogues based on video annotations, involving the following steps:

\begin{figure*}[!t]
\centering
\includegraphics[width=0.95\linewidth]{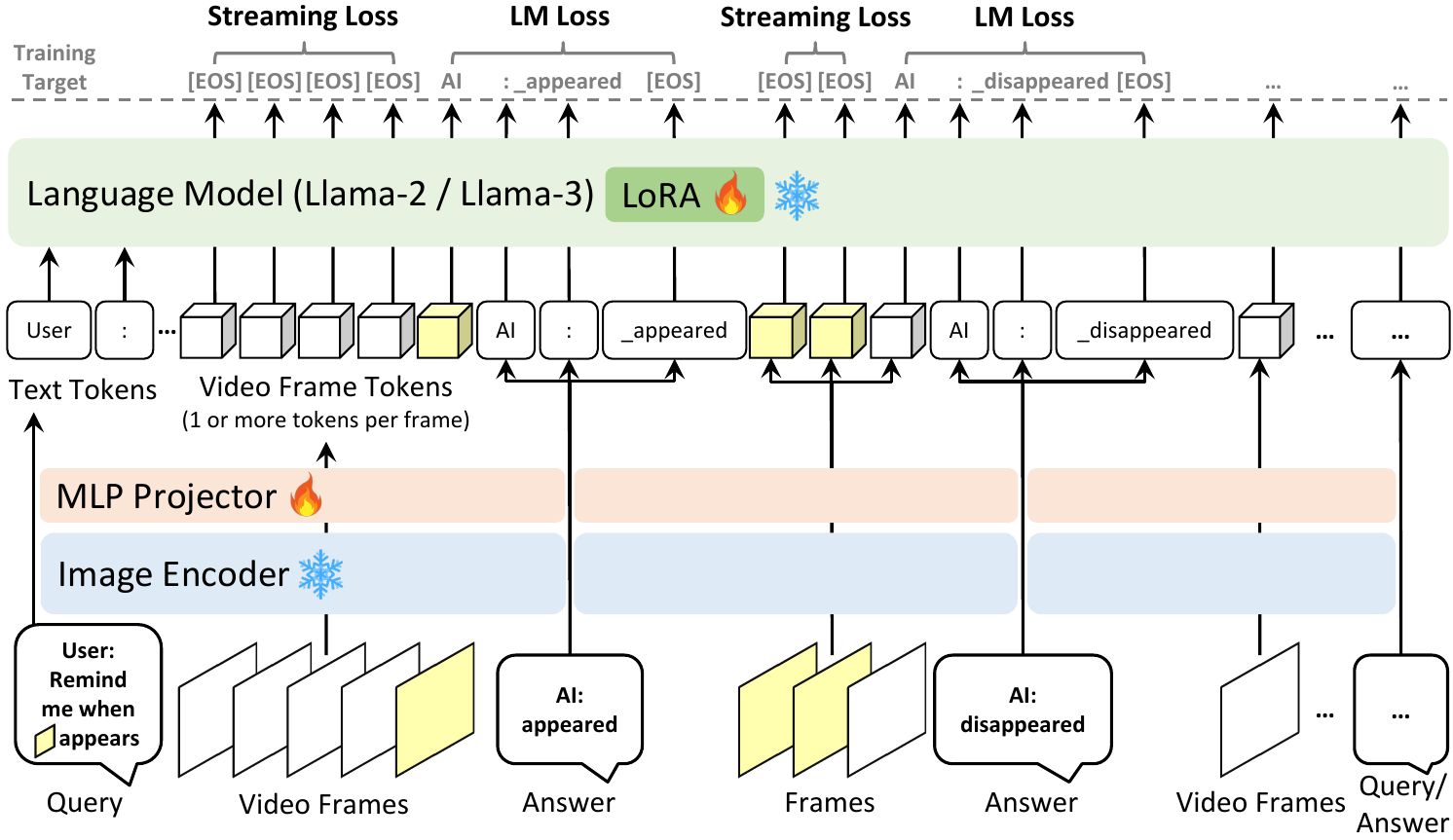}
\caption{\textbf{The training method in our LIVE framework}. We organize the user-assistant dialogue data and video frames in temporal order as the input sequence. To learn the model when to answer or keep silent in a video stream, we employ not only the standard language modeling (LM) loss but also introduce a streaming EOS prediction loss. This additional loss supervises the model when it is necessary to generate language, enabling it to produce temporally aligned responses and reduces the redundant dialogue history.}\label{figure:model}
\end{figure*}

\begin{itemize}
    \item First, we prepare a question template library containing various queries about the past, present, and future tenses of the video, totaling $N$ queries. We randomly sample one question from the library, denoted as $Q_i$.
    
    \item Then, we obtain the video annotation timeline from the offline dataset. This usually includes timestamped language descriptions, which we organize into a language prompt, e.g., ``time $t_a \sim t_b$: boiling the water; time $t_c \sim t_d$: cutting the vegetables.'', denoted as $A$. We consider all the state change critical timestamps as the ideal response times. For this example, $t_a, t_b, t_c,$ and $t_d$ are all considered response times.

    \item Third, we prompt the large language model to generate responses at every critical timestamp, \eg, $t_a, t_b, t_c, t_d$, according to $Q_i$ and $A$. We can repeat this procedure for each $Q_i$ until all queries have been processed. The responses are saved for loading during training.
    
    \item Finally, during training, we (1) randomly sample a query and load its responses at critical timestamps, (2) randomly insert a query into a video timestamp $t_r$, (3) discard the responses that occur before $t_r$, and add a response at $t_r$. Here different queries can be inserted into one video, which only requires discarding the responses of the previous query after the new query insertion timestamp.
\end{itemize}

In this way, we can generate temporally varied and free-form dialogue data within a video stream. We have prepared 50 questions each for past, current, and future events, totaling $N=150$ queries. We use Llama-2-13B-Chat~\cite{llama2} or Llama-3-8B-Instruct~\cite{llama3} to generate the responses and insert a maximum of 3 queries per training sample. The offline datasets we used are COIN~\cite{coin} and Ego4D GoalStep~\cite{ego4d_goalstep} (for demo usage), which belong to the categories of egocentric and instructional video datasets, aligning with our aim to develop online video assistants. Here, we do not consider online action detection benchmarks (e.g., THUMOS14~\cite{thumos14}, TVSeries~\cite{oad}) because they are closed-set online classification benchmarks, and their labels are too brief, which may cause language models to generate hallucinatory responses. Please refer to the supplementary material for the generated dialogue example.

\subsection{Model Training}

\noindent\textbf{Model Architecture.}
We illustrate the model architecture in Figure~\ref{figure:model}. Similar to LLaVA~\cite{llava,llava1.5}, it comprises three key components: an image encoder, an MLP projector, and a language model. For the image encoder, we utilize the CLIP ViT-L~\cite{clip,vit} encoder (pretrained on DataComp-1B~\cite{datacomp}) to extract video frame embeddings at 2 FPS. Each video frame embedding has a shape of $(1+h_p\times w_p)\times c$, where $(1+h_p\times w_p)$ denotes the CLS token and average pooled spatial tokens.\footnote{The experiments described in this paper are conducted without extra spatial tokens (\ie, $h_p = w_p = 0$), which is the most efficient setup and can handle half-hour videos within a 4096 context window. Our released models for demo usage include $1+h_p\times w_p = 1+3\times3=10$ tokens, offering better detail in dialogue but supporting shorter maximum video lengths. Despite more tokens used per frame, all models can still run at over 10 FPS for 5-minute Ego4D narration streams.} The extracted frame embeddings from the image encoder are then fed into MLP projector to frame tokens, as in LLaVA-1.5~\cite{llava1.5}. Then frame tokens are interleaved with language tokens as input to an LLM, Llama-2-7B-Chat~\cite{llama2} or Llama-3-8B-Instruct~\cite{llama3}. Finally, we incorporate LoRA~\cite{lora} in every linear layer of the LLM for efficient tuning. 

\noindent\textbf{Training Loss.}
As described in Section~\ref{section:video_streaming_dialogue}, our learning objective is twofold. The first part focuses on auto-regressive language modeling, aiming to maximize the joint probability of input text sequences. The second training objective involves streaming EOS prediction, which requires the model to remain silent when it is unnecessary to output responses. With these two training objectives, we have language modeling (LM) loss and streaming loss terms to minimize, both employing cross-entropy loss:

\begin{equation}
 L = \frac{1}{N}\sum^N_{j=1}(\underbrace{-\log l_{j+1}P_j^\texttt{[Txt$_{j+1}$]}}_{LM Loss} - \underbrace{w\log f_jP_j^\texttt{[EOS]}}_{Streaming Loss}),
\end{equation}

\noindent where $l_j$ and $f_j$ denote condition indicators. $l_j$ is 1 if the $j$-th token is a language response token, and 0 otherwise. $f_j$ is 1 if (1) the $j$-th token is the \textit{last} token of a frame\footnote{When a frame has multiple patch tokens, loss is only on the last one.}, and (2) $l_{j+1} = 0$. In essence, the streaming EOS loss is applied to frames before responding. $P_j^\texttt{[Txt$_{j+1}$]}$ denotes the probability on $j+1$-th text token, output from the language model head of the $j$-th token, and $P_j^\texttt{[EOS]}$ represents that probability for the EOS token. $w$ is a balance term, set to $w = 1$ by default. As shown in Figuer~\ref{figure:model}, we visualize the ranges of language loss and streaming loss in an input sequence when we only use 1 token for each frame.

\begin{figure}[t]
\centering
\includegraphics[width=\linewidth]{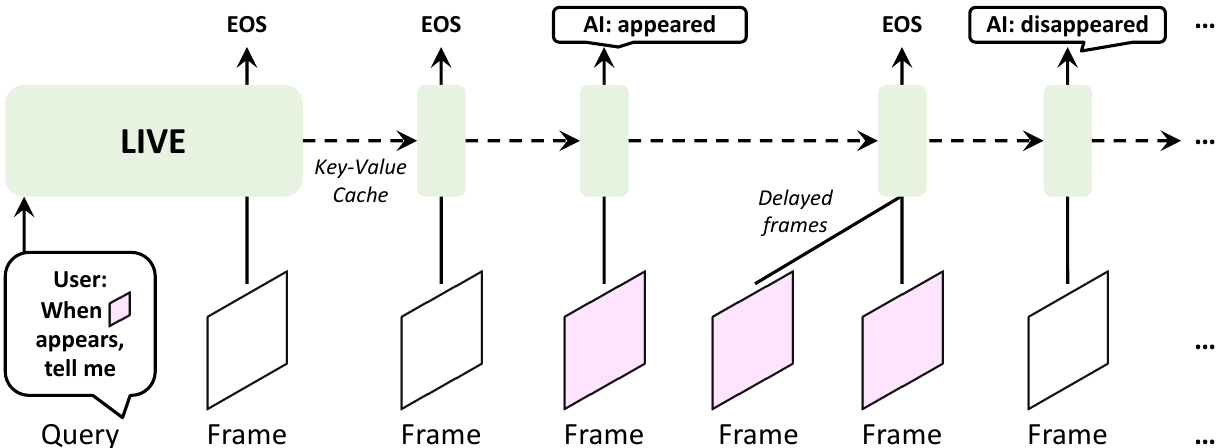}
\caption{\textbf{Inference pipeline in our LIVE framework}. During inference, video frames serve as streaming inputs. Our model maintains a continuous key-value cache as the input progresses to speed up the inference. Furthermore, we parallelize the fast video frame encoder and the slower language model to avoid the bottleneck in the latter. Video frame tokens can be always encoded and buffered, no need to wait the language decoding.}\label{figure:inference}
\end{figure}

\subsection{Inference}

\noindent\textbf{Probability Correction.} The prevalence of EOS token will bias the model towards EOS token prediction. To address this, we introduce a threshold $\theta$ to correct the output probability on frame tokens: EOS will not be considered as the next token if $P_j^\texttt{[EOS]} < \theta$. In practical usage, we find that setting $\theta$ to 0.5$\sim$0.8 yields much better results than no threshold here.

\noindent\textbf{Continuous Key-Value Cache.} As shown in Figure~\ref{figure:inference}, During inference, the video is input as a frame-by-frame stream, with a default FPS 2. Our model takes the current frame as input and generates tokens on-the-fly. In the whole process, we use the key-value cache trick to accelerate token decoding, thus we do not need to manually append the generated tokens for the next frame. As our training encourages the model to keep silence, this continuous inference would be efficient, providing the possibility to pace with the video stream speed.

\noindent\textbf{Parallelization of encoding and decoding.} Our video frame encoder utilizes CLIP ViT-L (307M), which is significantly smaller than the 7B/8B LLM. This size discrepancy leads to a speed mismatch, potentially resulting in frame skipping when the LLM decodes long sentences. To mitigate this issue, we parallelize the processes and establish a FIFO queue for video frame tokens. The fast encoder does not need to wait the slow LLM, it just always encode the video frames and append them to the queue. Once the language model completes its previous frame decoding, it can fetch the frame tokens in the queue but do not delay the video encoding.
\section{Experiments}

\subsection{Implementation Details}

We implement VideoLLM-online with our LIVE framework. It has two versions: 

$\bullet$ The more efficient one, VideoLLM-online-7B-v1, using OpenCLIP-ViT-L-224~\cite{clip,datacomp} as the video frame encoder, a 2-layer MLP as the connector, and Llama-2-7B-Chat~\cite{llama2} as the language model. Each video frame only costs 1 CLS token.

$\bullet$ The more effective one, VideoLLM-online-8B-v1+, using SigLIP-ViT-L-384~\cite{siglip} as the video frame encoder, a 2-layer MLP as the connector, and Llama-3-8B-Instruct~\cite{llama3} as the language model. Each video frame costs 1 CLS token and 3$\times$3 tokens by average pooling, \ie, 10 tokens per frame.

By default, the experiments in this paper are conducted with VideoLLM-online-7B-v1 due to our limited computation resources. We also train a VideoLLM-online-8B-v1+ model for demo purpose.

For model training, all models are trained with with 2 FPS sampled videos.\footnote{The inference can be set with higher FPS. We tried 10 FPS in some examples and we do not observe obvious performance degradation when inference FPS is 10.} We train model with LoRA~\cite{lora} to all linear layers, with a rank of 128, scaling factor of 256. For the sake of simplicity, the training is directly performed without vision-language aligning procedure~\cite{llava}. We also tried to use LlaVA-1.5~\cite{llava1.5} to initialize our connector and LLM, but we found the performance is similar, thus we just keep the MLP randomly initialized. For video streaming dialogue experiments, we train 2 epochs for the model. For the downstreaming offline experiments, we train 5$\sim$6 epochs without pre-training for fair comparison with previous methods. By default, we set streaming loss weight $w=1.0$ during training.

\subsection{Evaluation Setting}

\textbf{Datasets.} We use (1) assistance-related, instructional video dataset COIN~\cite{coin} and (2) continuous, egocentric video dataset Ego4D~\cite{ego4d} in various settings:
\begin{itemize}
    \item \textbf{Ego4D Narration Stream}: We also leverage the dense Ego4D timestamp-narration to create a streaming set. The goal is to generate narrations timely like Ego4D human annotators~\cite{ego4d}. We follow the division of the training, validation, and test set in EgoVLP~\cite{egovlp}. 
    \item \textbf{COIN+Ego4D Narration Stream}: To further evaluate the potential of model's performance to free-form dialogue, we construct a simple COIN+Ego4D Stream set, constructed from COIN annotations using our data generation methods, and the above Ego4D Narration Stream. The model should remind the user when an action starts, summarizes the action when it ends, as well as forecasts the next action. We use the same training/testing splits as COIN benchmarks. See appendix for details.
    \item \textbf{Ego4D GoalStep+Narration Stream}: Due to potential privacy risks associated with COIN dataset collected from YouTube videos, we have opted to use Ego4D GoalStep~\cite{ego4d_goalstep} for training our released model. While it is also capable of online chatting, it may exhibit limitations when dealing with third-person perspective videos.
    \item \textbf{COIN Benchmarks}: Following on previous studies~\cite{distantsup,procedurevrl,videotf,unloc}, we evaluate our model on six common benchmarks of the COIN dataset: step recognition, step forecasting, task summarization, procedure forecasting, procedure forecasting with a goal.  
    \item \textbf{Ego4D long-term action anticipation (LTA) benchmark}: This benchmark requires to predict next $Z = 20$ actions (verbs and nouns) for the given video of previous 8 steps. We use the standard Ego4d v2 splits as in previous studies~\cite{antgpt,palm}. 
\end{itemize}

\noindent\textbf{Evaluation metrics.} We use the following metrics to evaluate the model as an online video assistant:

\begin{itemize}
\item \textbf{Language Modeling Metrics.} We use common language perplexity to indicate the language modeling capability (\textit{LM-PPL}) at a given timestamp. A lower \textit{LM-PPL} signifies better accuracy in answering. However, this metric is not suitable for comparing different LLMs due to potential variations in language tokenization. Therefore, we calculate the language generation matching ratio (\textit{LG-Match}) to compare VideoLLM-online-7B-v1 and VideoLLM-online-8B-v1+. Note that \textit{LG-Match} is calculated in an autoregressive order, meaning it represents the ratio of the position of the first error token to the total number of tokens.

\item \textbf{Time Difference (TimeDiff).} To evaluate the temporal alignment capability of an online assistant, we calculate the discrepancy between the timestamp of its response and the expected timestamp for each response. We average TimeDiff each turn as the metric.

\item \textbf{Fluency.} Individual \textit{LM-PPL}, \textit{LG-Match} or \textit{TimeDiff} do not entirely evaluate both language and temporal effectiveness in a streaming dialogue. We introduce the Fluency metric, which evaluates the proportion of consecutive successful token prediction within a dialogue turn. As the token also including language tokens, Fluency can comprehensively reflect the language modeling in an online streaming.
\end{itemize}

We would like to note that these metrics are mainly for monitoring model performance in the streaming narration task. Firstly, the narration text is relatively simple, mainly composed of a subject, verb, and object, thus the \textit{LM-PPL} and \textit{LG-Match} can still work for this simple language. Secondly, the streaming narration requires the human annotator to write the description at the moment when the action state has changed, which is very suitable for validating the time differences. However, these metrics are not so effective for evaluating more complicated, free-form online conversation scenarios, and this is also a common problem in evaluating free-form LLM generation. We leave this for future work.

\noindent \textbf{Baselines.} 
To our best knowledge, we are the first one to tackle producing temporal aligned, free-form language answer with streaming video settings. To better understand the challenges, we build baseline models for video-text interleaved dialogue, per-frame dialogue, as we described in Section~\ref{section:video_streaming_dialogue}, with the same model architecture and training details to VideoLLM-online, differing only in their training objective and multi-turn formulation. 

\begin{table*}[t]
\scriptsize
\centering
\begin{subtable}[h]{1.0\textwidth}
\centering
\begin{tabular}{l|l|ccccc}
\toprule
\multirow{2}{*}{Method} & \multirow{2}{*}{Training Objective} & \multicolumn{3}{c}{Ego4D Narration Stream on Validation} & \multirow{2}{*}{\#Training Token$\downarrow$} & \multirow{2}{*}{Training Cost} \\
&  & \textit{LM-PPL}$\downarrow$   & \textit{TimeDiff}$\downarrow$ &  \textit{Fluency}$\uparrow$ \\
\midrule
\multicolumn{2}{c|}{No Training} & 498.5 & 6.50 & 0.1\% & n/a & n/a \\
\midrule
Interleaved Dialogue  & Language Modeling & 2.45 & 6.47 & 11.1\% & \textbf{1694} & \textbf{12h} \\
\midrule
Per-frame Dialogue for Streaming  & Language Modeling (w/ EOS turns) & 3.34 & 2.52 & 37.7 \% & 6737 & 22h \\
\midrule
\baseline{Streaming Dialogue (Ours)}& \baseline{Language Modeling + Streaming EOS} & \baseline{\textbf{2.43}}  & \baseline{\textbf{2.32}} &  \baseline{\textbf{42.6\%}}  &  \baseline{\textbf{1694}} & \baseline{\textbf{12h}} \\
\bottomrule
\end{tabular}
\caption{\textbf{Learning method for streaming dialogue}. Training with streaming dialogue method can achieve much better \textit{TimeDiff} and \textit{Fluency}, as well as maintain the language modeling quality. Meanwhile, the streaming dialogue can enjoy much more efficient training than per-frame dialogue for video streaming dialogue.}
\label{tab:ablation_learning}
\end{subtable}

\begin{subtable}[h]{0.4\textwidth}
\centering
\begin{tabular}{l|ccc}
\toprule
\multirow{2}{*}{Streaming Loss} & \multicolumn{3}{c}{Ego4D Narration Stream Validation} \\
&\textit{LM-PPL}$\downarrow$   & \textit{TimeDiff}$\downarrow$ &  \textit{Fluency}$\uparrow$ \\
\midrule
\baseline{Standard CE} & \baseline{\textbf{2.43}}  & \baseline{\textbf{2.32}} &  \baseline{\textbf{42.6\%}} \\
OHEM~\cite{ohem} & 2.53 & 2.39 & 41.0\% \\ 
Focal Loss~\cite{focal_loss} & 2.59 & 2.44 & 39.4\% \\
\bottomrule
\end{tabular}
\caption{\textbf{Streaming loss function}. Standard CE (cross-entropy) is enough for training streaming dialogue; there is no need to specifically to address the class imbalance on EOS token.}
\label{tab:ablation_stream_loss_func}
\end{subtable}
\begin{subtable}[h]{0.37\textwidth}
\centering
\begin{tabular}{l|ccc}
\toprule
\multirow{2}{*}{Weight $\tau$} & \multicolumn{3}{c}{Ego4D Narration Stream Validation} \\
&\textit{LM-PPL}$\downarrow$   & \textit{TimeDiff}$\downarrow$ &  \textit{Fluency}$\uparrow$ \\
\midrule
$\tau = 0.5$ & 2.44 & 2.32 & 42.4\% \\
\baseline{$\tau = 1.0$} & \baseline{\textbf{2.43}}  & \baseline{2.32} &  \baseline{\textbf{42.6\%}} \\
$\tau = 2.0$ & 2.46 & \textbf{2.31} & 42.5\% \\
$\tau = 3.0$ & 2.47 & 2.32 & 42.5\% \\
\bottomrule
\end{tabular}
\caption{\textbf{Streaming loss weight}. Using slightly higher streaming loss weight ($\tau = 2.0$) achieves the best trade-off among various metrics.}
\label{tab:ablation_stream_loss_weight}
\end{subtable}
\begin{subtable}[h]{0.22\textwidth}
\centering
\begin{tabular}{c|cc}
\toprule
Method &\textit{Mem}$\downarrow$ & \textit{FPS}$\uparrow$  \\
\midrule
Interleaved & 34.4G & 1.5 \\
\midrule
\multirow{2}{*}{\shortstack{Per-frame\\Streaming}} & \multirow{2}{*}{24.9G} & \multirow{2}{*}{7.5} \\
 & & \\
\midrule
\baseline{Streaming} & \baseline{\textbf{18.2G}} & \baseline{\textbf{13.5}} \\
\bottomrule
\end{tabular}
\caption{\textbf{Generation memory/speed}. Streaming dialogue method has much better efficiency.}
\label{tab:ablation_inference}
\end{subtable}
\caption{\textbf{Ablation experiments on Ego4D Narration Stream}. We train VideoLLM-online on Ego4D~\cite{ego4d} narration stream training set and evaluate on its validation set. The comparison is based on our designed metrics: the ratio of strictly correct prediction tokens (\textit{Fluency}), language modeling perplexity (\textit{LM-PPL}) and time difference (\textit{TimeDiff}) metrics. \#Training Token denotes the average token length during training. \textit{TimeDiff} refers to difference in second. \colorbox{verylightblue}{Default settings} are highlighted.}
\end{table*}

\subsection{Ablation Study}

\noindent\textbf{Learning Method}.
Table~\ref{tab:ablation_learning} shows the ablation studies on learning methods in a streaming setting. Both vision-language interleaved and streaming methods exhibit low perplexity loss, indicating that our proposed objective does not hurt language modeling capability. However, learning with per-frame for streaming will produce significant higher \textit{LM-PPL} than others, which might be attributed to the too more single EOS token in answering that affects the original language modeling. 

When we turn to online metrics of \textit{TimeDiff} and \textit{Fluency}, streaming dialogue method yields much better results than others. In our observation, the first interleaved dialogue method always outputs language after every frame, and the second mutli-turn for streaming dialogue approach  tends to answer EOS token after every frame, which decreases their performance for streaming video inputs. Furthermore, per-frame for streaming dialogue method will significanly slow down the training speed due to its lengthy prompts, while our method has no negative impact on the efficiency.

\noindent\textbf{Streaming Loss}.
We continue to investigate the most suitable strategy to learn the streaming objective. As shown in Table~\ref{tab:ablation_stream_loss_func} and Table~\ref{tab:ablation_stream_loss_weight}, we find a default setting works surprisingly well (CE loss, $\tau=1.0$), which demonstrates there is no need to apply more advanced loss (\eg Focal Loss~\cite{focal_loss}) to address the imbalance on EOS token.

\noindent\textbf{Inference Efficiency}. In Table~\ref{tab:ablation_inference}, we test the inference efficiency on Ego4D narration stream validation set (5 minute), and report the memory cost and average FPS on a single A100 GPU. The first interleaved dialogue method, which will output language after every video frame, has huge memory cost and slow generation speed. The second one, per-frame dialogue for streaming that formulates all in a multi-turn dialogue, show better efficiency than the first one since it can cost less tokens in redundant frames. However, this approach still lags significantly behind our streaming dialogue approach, which does not cost extra tokens in redundant frames thus maintain smaller key-value cache. We observe for most Ego4D  videos, our model can run larger than 10 FPS, providing possibility for AI assistants working in real-time video stream.

\begin{table*}[t]
\centering
\scriptsize

\begin{subtable}[h]{0.53\textwidth}
\scriptsize
\centering
\begin{tabular}{l|c|ccccc}
\toprule
\multirow{2}{*}{Method} & \multirow{2}{*}{\shortstack{Not use\\HT100M}} & \multicolumn{5}{c}{COIN Benchmark Top-1 Accuracy$\uparrow$} \\
& & Step & Task & Next & Proc. & Proc.+ \\
\midrule
ClipBERT~\cite{clipbert}  & \checkmark & 30.8 & 65.4 & - & - & - \\
TimeSformer~\cite{timesformer}  & \xmark & 46.5 & 85.3 & 34.0 & 17.0 & 40.1 \\
Paprika~\cite{paprika}   & \xmark & 51.0 & 85.8 & 43.2 & - & - \\
DistantSup~\cite{distantsup}   & \xmark & 54.1 & 90.0 & 39.4 & - & 41.3 \\
VideoTF~\cite{videotf} & \xmark & 56.5 & 91.0 & 42.4 & 40.2 & 46.4 \\
ProcedureVRL~\cite{procedurevrl}  & \xmark & 56.9 & 90.8 & 46.8 & - & - \\
VideoTaskGraph~\cite{video_mined_task_graph} & \xmark & 57.2 & 90.5 & 40.2 & - & - \\
\baseline{VideoLLM-online-7B-v1} & \baseline{\checkmark} & \baseline{59.8} & \baseline{92.1} & \baseline{48.1} & \baseline{47.9} & \baseline{52.9} \\
\baseline{VideoLLM-online-8B-v1+} & \baseline{\checkmark} & \baseline{\textbf{63.1}} & \baseline{\textbf{92.7}} & \baseline{\textbf{49.1}} & \baseline{\textbf{49.8}} & \baseline{\textbf{54.1}} \\ 
\bottomrule
\end{tabular}
\caption{Results on COIN benchmarks (left to right): step recognition, task recognition, next forecasting, procedure forecasting, procedure forecasting with a goal.}
\end{subtable}
\begin{subtable}[h]{0.46\textwidth}
\scriptsize
\centering
\begin{tabular}{l|l|l|ccc}
\toprule
\multirow{2}{*}{Method} & \multirow{2}{*}{\shortstack{Not use\\EgoVLP}} & \multirow{2}{*}{\shortstack{End-to\\-end?}} & \multicolumn{3}{c}{Ego4D LTA ED@Z=20$\downarrow$} \\
& & & Verb & Noun & Action \\
\midrule
CLIP~\cite{video+clip4lta} & \checkmark & \checkmark &0.739&0.769&0.941\\
EgoT2~\cite{egot2} & \checkmark &\checkmark&0.722&0.764&0.935 \\
I-CVAE~\cite{icvae} &\checkmark & \checkmark &0.753&0.749&0.931\\
HierVL~\cite{hiervl} &\checkmark & \checkmark & 0.724&0.735&0.928\\
VideoLLM~\cite{videollm}  & \xmark & \checkmark & 0.721 & 0.725 & 0.921 \\
\baseline{VideoLLM-online-7B-v1} & \baseline{\checkmark} & \baseline{\checkmark} & \baseline{0.697} & \baseline{0.698} & \baseline{0.897} \\
\baseline{VideoLLM-online-8B-v1+} & \baseline{\checkmark} & \baseline{\checkmark} & \baseline{\textbf{0.689}} & \baseline{\textbf{0.671}} & \baseline{\textbf{0.884} }\\
\midrule
\textcolor{gray}{Palm~\cite{palm}} & \xmark & \xmark  &\textcolor{gray}{0.696} &\textcolor{gray}{0.651} & \textcolor{gray}{0.886}\\
\textcolor{gray}{AntGPT~\cite{antgpt}} & \xmark & \xmark &\textcolor{gray}{0.650}&\textcolor{gray}{0.650}&\textcolor{gray}{0.877}\\
\end{tabular}
\caption{Results on Ego4D LTA benchmark, evaluated on \href{https://eval.ai/web/challenges/challenge-page/1598/evaluation}{public server}. ED@Z=20 denotes editing distance for future 20 actions.}
\end{subtable}

\caption{Experiments on COIN~\cite{coin} and Ego4D~\cite{ego4d} benchmarks. VideoLLM-online is finetuned on their training set, and strictly evaluated on the test set by generated string comparison with the ground-truth text. It achieves best results among end-to-end models.\vspace{-3mm}}\label{exp:results}
\label{table:coin_ego4d_benchmark}
\end{table*}

\subsection{Results}

\noindent\textbf{Offline Language Modeling}. We show our model can perform well on traditional temporal summarization and forecasting problems. As shown in Table~\ref{table:coin_ego4d_benchmark}(a), our model achieves state-of-the-art performances in  step/task summarization and next step/procedure forecasting benchmarks of COIN dataset~\cite{coin}. Furthermore, we also obtain the best performance among end-to-end models evaluated on Ego4D LTA. Although the results of AntGPT~\cite{antgpt} are better than us, they used egocentric pre-trained visual feature~\cite{egovlp}, and integrates lots of complex cascading methods to improve the forecasting results. Our VideoLLM-online, however, directly outputs language as the results, which performs better than the similar end-to-end VideoLLM~\cite{videollm}. 

\begin{table}[h]
\centering
\scriptsize
\begin{tabular}{l|ccc}
\toprule
\multirow{2}{*}{Method} & \multicolumn{3}{c}{Ego4D Narration Stream Validation} \\
&  \textit{LG-Match}$\uparrow$ & \textit{TimeDiff}$\downarrow$ &  \textit{Fluency}$\uparrow$  \\
\midrule
VideoLLM-online-7B-v1 & 42.3\% & 2.25 & 42.6\% \\ 
VideoLLM-online-8B-v1 & 48.3\% & 2.05 & 45.2\%  \\
VideoLLM-online-8B-v1+ & \textbf{49.0\%} & \textbf{2.05} & \textbf{45.3\%}  \\
\bottomrule
\end{tabular}
\caption{Performance comparison of VideoLLM-online variants.}\label{table:ego4d_stream_llama2/3}
\end{table}

\noindent\textbf{Comparison between Model Variants.} We compare VideoLLM-online-7B-v1, VideoLLM-online-8B-v1, and VideoLLM-online-8B-v1+ on the Ego4D narration stream task. ``7B'' and ``8B'' refer to Llama-2-7B and Llama-3-8B, respectively, while ``v1'' and ``v1+'' indicate the usage of either one token per frame or multiple tokens per frame. As shown in Table~\ref{table:ego4d_stream_llama2/3}, the enhanced language model significantly improves performance across all aspects. Utilizing more tokens per frame enhances the vision-language capability, albeit with limited benefits to online performance.

\noindent\textbf{Visualization}. In Figure~\ref{figure:teaser}, we visualize two representative examples, real-time narration and online dialogue. The most distinctive characteristics of our approach are: (1) the dialogue process goes along with the streaming video input, rather than chatting based on the full video. (2) The response will be ``muted'' when it is unnecessary, significantly improving the overall speed of video streaming dialogue. 

Another example is shown in Figure~\ref{figure:align}. It can be seen that our model demonstrates strong alignment between the streaming visual frames and the output responses. With our efficient inference strategy, we can envision an J.A.R.V.I.S-like intelligent assistant that can assist users in real time.
\section{Conclusion}
We propose Learning-In-Video-strEam (LIVE), a novel framework empowering LLMs to handle streaming video, to produce temporal aligned answers, hold long-context video duration, and have high inference efficiency. We use LIVE to train a simple VideoLLM-online model, which not only achieves superior capability in online/offline vision-language tasks, but also enable fast inference for an online video streaming setting. We believe enabling such abilities will be an important step to move towards always-on online assistant. In future work, to make our VideoLLM-online be more general and improve its spatial capability in zero-shot prediction for downstream applications, we will explore suitable pre-training data source, and develop models that can employ more spatial tokens but without many trade-off on speed and memory cost.

\quad

\noindent\textbf{\large{Acknowledgment}}
This work is sponsored by Project Aria Team, Meta. The datasets and processing were acquired and all models were trained at the National University of Singapore (NUS) by NUS authors.

{
    \small
    \bibliographystyle{ieeenat_fullname}
    \bibliography{ai}
}

\clearpage
\setcounter{page}{1}
\maketitlesupplementary
\renewcommand{\thesection}{\Alph{section}}
\setcounter{section}{0} 
\noindent This supplementary material includes following sections:
\begin{itemize}
    \item Section~\ref{section_supp:comparison} provides an analysis of per-frame chatting. More specifically, we prompt GPT-4V for video streaming dialogue and compare it with interleaved vision-language dialogue and our method.
    \item Section~\ref{section_supp:data} elaborates on data details, especially on prompts, including examples of Ego4D Narration Stream, COIN Dialogue Stream, training and inference prompts, and evaluation schemes for COIN benchmarks and Ego4D LTA.
    \item Section~\ref{section_supp:more_reslts} shows the results on Ego4D+COIN stream set. Meanwhile, we show some demo results from the VideoLLM-online-model.
    \item Section~\ref{section_supp:limit} discusses some limitations of the paper.
    Please also refer to our released repository at \href{https://showlab.github.io/videollm-online}{showlab.github.io/videollm-online} for more implementation details.
\end{itemize}

\begin{figure*}
    \centering
    \includegraphics[width=\linewidth]{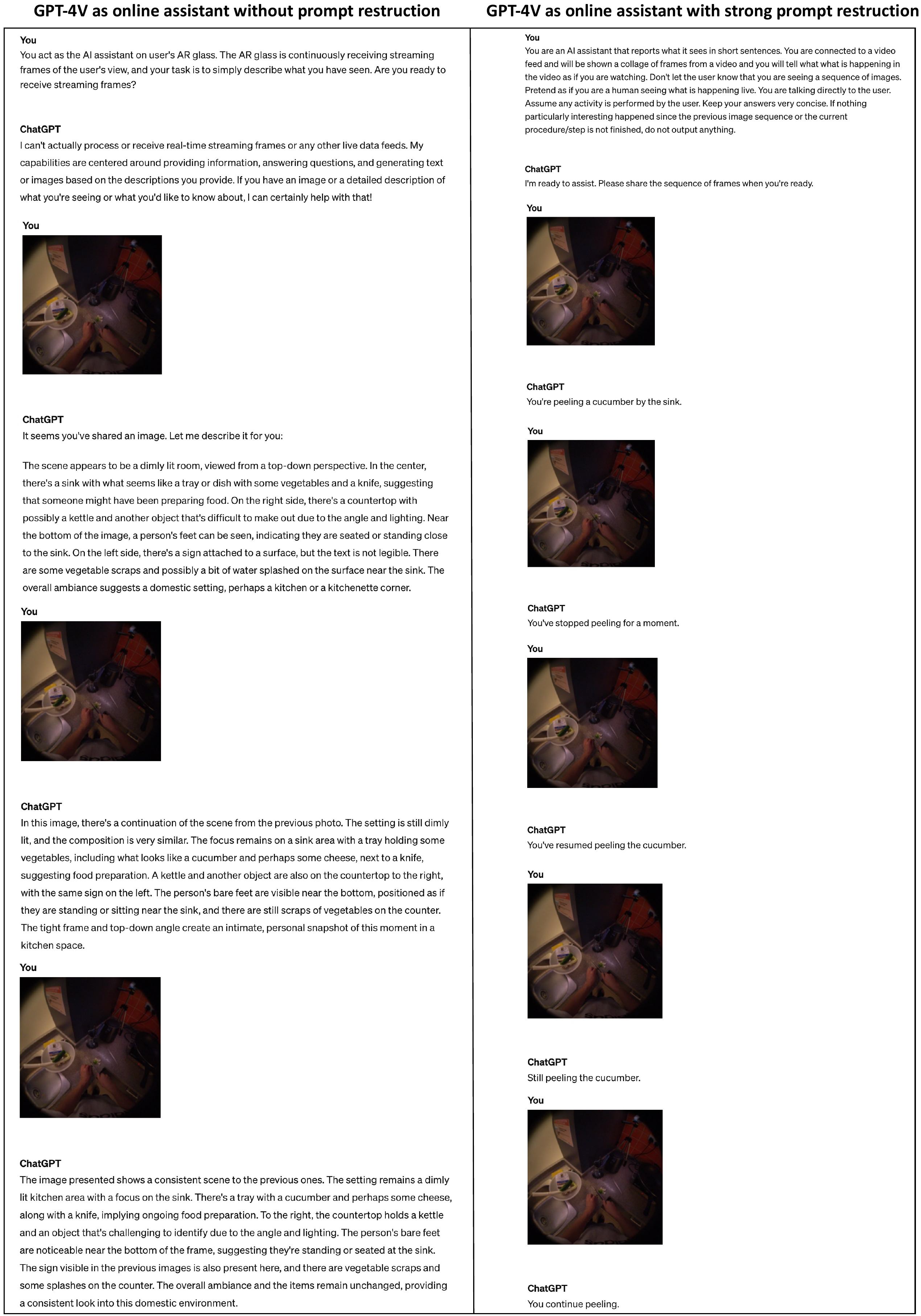}
    \caption{GPT-4V examples to real-time narration, with no prompting restriction (left) and strong prompting restriction (right). }
    \label{figure_supp:gpt4v}
\end{figure*}

\section{Analysis to Per-frame Chatting}\label{section_supp:comparison}

As shown in Figure~\ref{figure_supp:gpt4v}, we prompt GPT-4V to do the real-time narration task. In ideal case, we hope the model just output narration like ``cutting vegetables'' at the first frame, since these frames are nearly no change. We use two methods of prompting: (1) no prompting restriction: this prompt allows the GPT-4V to output language at every frame, without consideration on the conciseness. See Figure~\ref{figure_supp:gpt4v} left part, we can observe that the response of GPT-4V is very lengthy, making it impossible for real-time usage; (2) with strong prompting restriction: the right part of the figure suggests that GPT-4V can be prompted to approach the video streaming dialogue. However, it is still per-frame dialogue and still cost tokens and times per frame. Moreover, we find it is not so stable; sometimes there would be obvious hallucination that may not be appeared in GPT-4V level, like ``you are peeling'' vs. ``you have stopped'' at the first and second frame.



\section{More Data Details}\label{section_supp:data}

\subsection{Data Construction}

\noindent\textbf{COIN Stream Set.} 
This set is derived from COIN annotations, adapted using our streaming dialogue generation schemes. Initially, a user query outlines the video's overall task, prompting the model to track and record the activities shown. The model is then required to watch the video and provide real-time responses. An example of this process is provided in Section~\ref{sec_supp:prompt}. It's important to note that this dataset for experiment has a relatively fixed structure for stable evaluation, \ie, the user query occurs only at the beginning, which simplifies the evaluation process. However, the models use for demo, as depicted in Figure 1 of the paper, is trained with randomized queries, timestamps, and varying numbers of turns. 

\noindent\textbf{Ego4D Narration Stream Set.} 
The annotation process for Ego4D Narration inherently follows a streaming dialogue format. Initially, videos are segmented into clips, each with a maximum duration of five minutes, for the purpose of acquiring narrations. Annotators are then tasked with providing a concise summary narration, typically 1-3 sentences long, for each clip. Once they have established an overall understanding of the clip, they proceed to write detailed, play-by-play descriptions of the actions. Here we only use the second part, \ie the streaming narration. The training and inference prompts for Ego4D narration, which are adapted from the original text guidelines provided to annotators, are detailed in Section~\ref{sec_supp:prompt}. Note the narration for experiments are not refined by Llama; we use the original narration for stability (but remove special strings like ``\#C'', ``\#O'') .

\subsection{Training and Inference Prompt}\label{sec_supp:prompt}
\noindent\textbf{System Prompt.} We have a simple system prompt at the beginning of the dialogue:

\noindent\fbox{\parbox{\dimexpr\linewidth-2\fboxsep-2\fboxrule\relax}{\textit{A user wears AR glasses equipped with an intelligent assistant. The AR glasses continuously receive streaming video frames from the user's viewpoint, enabling the assistant to observe and provide real-time assistance in response to the user's queries when necessary. Below is their dialogue, accompanied by streaming video frames included in the user's query.}}}

\noindent In the following, we use \texttt{[System]} to denote it.

\noindent\textbf{Frame Placeholder.} 
In our training, each video frame is initially encoded using frozen CLIP ViT. These encoded frames are then projected into a language token-compatible space through a learnable MLP. We use \texttt{[F]} to denote tokens per frame. In our paper experiments, the number of tokens per frame is $|\texttt{[F]}| = 1$ for fast training/validation and parameters searching. For our demo, we use $|\texttt{[F]}| = 10$, \ie 1 CLS + $3\times3$ average pooled spatial token for better vision understanding ability.

\noindent\textbf{Streaming Dialogue Examples.} 
To illustrate the streaming dialogue format more clearly, we provide examples of training prompts from our generated COIN Dialogue Stream set and our curated Ego4D Narration Stream set. In these examples, tokens related to the streaming objective are highlighted in {\color{blue}blue}, while tokens associated with the language modeling objective are marked in {\color{orange}orange}. We ignored some chat  template strings (\eg, \texttt{[INST], [/INST]} in Llama~\cite{llama1,llama2}) for better visualization.

\begin{itemize}
    \item COIN Stream Example:

    \noindent\fbox{\parbox{\dimexpr\linewidth-2\fboxsep-2\fboxrule\relax}{
    
    \texttt{[System]}

    \textit{User: The video is about to install ceiling fan. Please remind me when the related action starts, summarizes when it ends, as well as forecasts the next action.}
    
    {\color{blue}\texttt{[F]}\texttt{[F]}...\texttt{[F]}}{\color{orange}\texttt{[F]}\textit{Assistant: Now doing the step to close switch. Then try to install fan tray.}}{\color{blue}\texttt{[F]}\texttt{[F]}...\texttt{[F]}}{\color{orange}\texttt{[F]}\textit{Assistant: Just finished the step to close switch. Then try to install fan tray.}}
    {\color{blue}\texttt{[F]}\texttt{[F]}...\texttt{[F]}}{\color{orange}\texttt{[F]}\textit{Assistant: Now doing the step to install fan tray. Then try to install fans and lights.}}
    {\color{blue}\texttt{[F]}\texttt{[F]}...}

    }}
    
    \item Ego4D Narration Dialogue Stream Example:

    \noindent\fbox{\parbox{\dimexpr\linewidth-2\fboxsep-2\fboxrule\relax}{
    
    \texttt{[System]}

    \textit{User: Please watch the video and narrate the video in real-time.}

    \color{blue}\texttt{[F]}\color{orange}{\texttt{[F]}\textit{Assistant: C walks around a room.}}\color{blue}\texttt{[F]}\texttt{[F]}\color{orange}{\texttt{[F]}\textit{Assistant: C picks up a wire from the floor.}}\color{blue}\texttt{[F]}\texttt{[F]}\texttt{[F]}\color{orange}{\texttt{[F]}\textit{Assistant: C pulls out a wire from a wall.}}\color{blue}\texttt{[F]}\texttt{[F]}\texttt{[F]}\texttt{[F]}\color{orange}{\texttt{[F]}\textit{Assistant: C looks around a room.}}\color{blue}\texttt{[F]}\texttt{[F]}...
    }}
\end{itemize}

\noindent\textbf{Benchmark Evaluation Prompt.} 
For benchmark evaluation, responses must adhere to a specific format. We incorporate an additional prompt in the user query for this purpose, denoted as \texttt{[BenchEval]}:

\noindent\fbox{\parbox{\dimexpr\linewidth-2\fboxsep-2\fboxrule\relax}{\textit{Please answer briefly for benchmark evaluation, and may use ; to separate different steps.
}}}

Additionally, we provide examples from all the benchmarks on which we have conducted evaluations:

\begin{itemize}
    \item COIN Step Recognition:

    \noindent\fbox{\parbox{\dimexpr\linewidth-2\fboxsep-2\fboxrule\relax}{\texttt{[System]}

    {\color{blue}\texttt{[F]}\texttt{[F]}$\cdots$\texttt{[F]}\texttt{[F]}}
    
    \textit{User: What was the previous step?}\texttt{[BenchEval]}
    
    \color{orange}\textit{Assistant: Take off the shell.}}
    }

    \item COIN Task Summarization:

    \noindent\fbox{\parbox{\dimexpr\linewidth-2\fboxsep-2\fboxrule\relax}{\texttt{[System]}

    {\color{blue}\texttt{[F]}\texttt{[F]}$\cdots$\texttt{[F]}\texttt{[F]}}
    
    \textit{User: What task can summarize these steps?}\texttt{[BenchEval]}
    
    \color{orange}\textit{Assistant: Cut and restore rope trick.}}
    }

    \item COIN Next Step Forecasting:

    \noindent\fbox{\parbox{\dimexpr\linewidth-2\fboxsep-2\fboxrule\relax}{\texttt{[System]}

    {\color{blue}\texttt{[F]}\texttt{[F]}$\cdots$\texttt{[F]}\texttt{[F]}}
    
    \textit{User: What is the next 1 step?}\texttt{[BenchEval]}
    
    \color{orange}{\textit{Assistant: Rotate body and accelerate the hammer.}} }
    }

    \item COIN Procedure Forecasting:

     \noindent\fbox{\parbox{\dimexpr\linewidth-2\fboxsep-2\fboxrule\relax}{\texttt{[System]}

    {\color{blue}\texttt{[F]}\texttt{[F]}$\cdots$\texttt{[F]}\texttt{[F]}}
    
    \textit{User: What are the next 5 steps?}\texttt{[BenchEval]}
    
    \color{orange}{\textit{Assistant: Insert it into the crystal head; fixe it with a crimping pliers; cut a certain length; insert it into the crystal head; fixe it with a crimping pliers.}}}
    }

    \item COIN Procedure Forecasting with Task Goal:

     \noindent\fbox{\parbox{\dimexpr\linewidth-2\fboxsep-2\fboxrule\relax}{\texttt{[System]}

    {\color{blue}\texttt{[F]}\texttt{[F]}$\cdots$\texttt{[F]}\texttt{[F]}}
    
    \textit{User: What are the next 2 steps to hang wallpaper?}\texttt{[BenchEval]}
    
    \color{orange}\textit{Assistant: Wipe or polish the wall; crop the wallpaper.}}
    }

    \item COIN Action Segmentation: 

     \noindent\fbox{\parbox{\dimexpr\linewidth-2\fboxsep-2\fboxrule\relax}{\texttt{[System]}

    \textit{User: Please output the corresponding action of each frame.}\texttt{[BenchEval]}
    
    {\color{blue}\texttt{[F]}\texttt{[F]}$\cdots$\texttt{[F]}}{\color{orange}\texttt{[F]}\textit{Assistant: Show the blank paper.}}{\color{orange}\texttt{[F]}\textit{Assistant: Show the blank paper.}}{\color{orange}\texttt{[F]}$\cdots$}{\color{blue}\texttt{[F]}\texttt{[F]}$\cdots$\texttt{[F]}}{\color{orange}\texttt{[F]}\textit{Assistant: Show the money to the audience.}}{\color{orange}\texttt{[F]}\textit{Assistant: Show the money to the audience.}\texttt{[F]}$\cdots$}{\color{blue}\texttt{[F]}\texttt{[F]}$\cdots$}
    }}

    \item Ego4D LTA:

     \noindent\fbox{\parbox{\dimexpr\linewidth-2\fboxsep-2\fboxrule\relax}{\texttt{[System]}

        {\color{blue}\texttt{[F]}\texttt{[F]}$\cdots$\texttt{[F]}\texttt{[F]}}
        
        \textit{User: What are the next 20 steps?}\texttt{[BenchEval]}
        
        \color{orange}{\textit{Assistant: apply flour; attach dough; knead dough; take dough; put dough; remove dough; knead dough; take dough; put dough; move dough; apply flour; knead dough; take dough; put dough; move table; apply flour; knead table; take dough; put dough; move dough.}}
        }}
    
\end{itemize}

\begin{figure*}[t]
    \centering
    \includegraphics[width=0.95\linewidth]{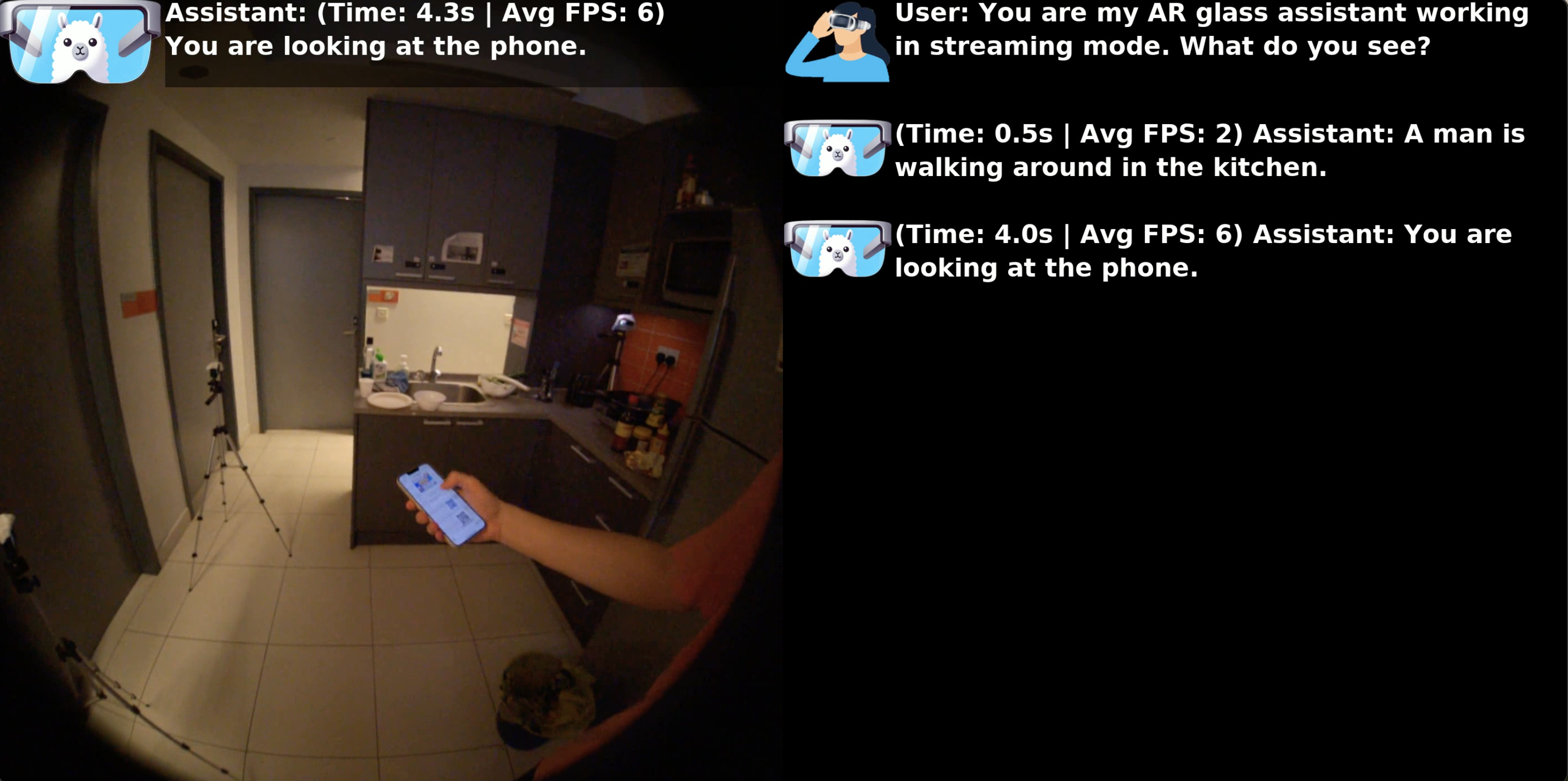}
    \caption{Online narration demo of VideoLLM-online.}
    \label{fig_supp:demo}
\end{figure*}

\begin{figure*}[t]
    \centering
    \includegraphics[width=0.95\linewidth]{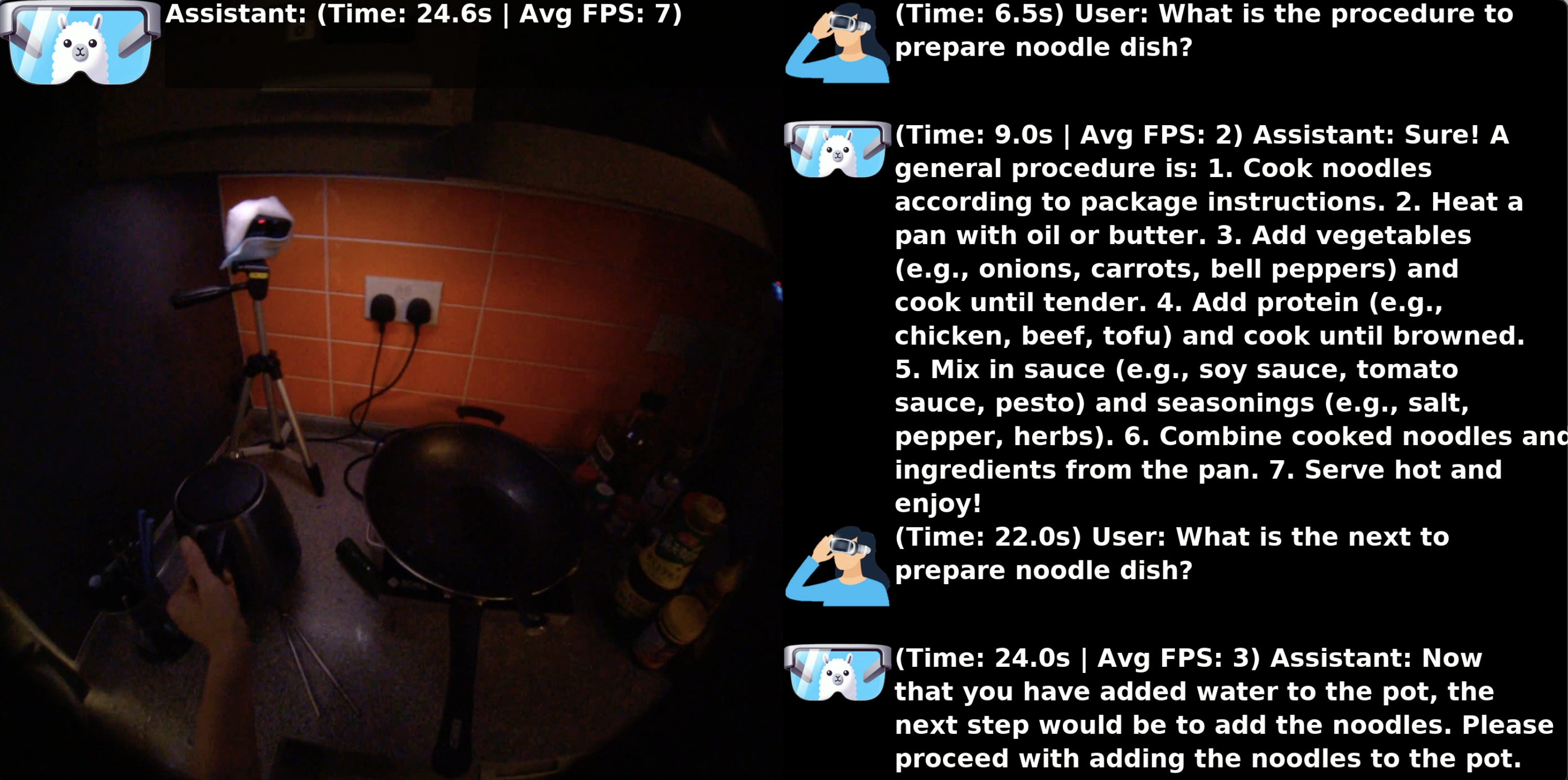}
    \caption{Online chatting demo of VideoLLM-online.}
    \label{fig_supp:demo}
\end{figure*}

\subsection{Evaluation Scheme}\label{sec_supp:eval}

We detail our methodology for evaluating performance on existing benchmarks.

\noindent\textbf{COIN Benchmarks.} 
Following the approach in~\cite{distantsup,procedurevrl,videotf}, we report top-1 accuracy for the COIN benchmarks. A unique challenge arises with Online-VideoLLM, as it produces outputs in natural language rather than class indices. To address this, we employ a simple string matching technique: we compare the model's language output with the COIN taxonomy dictionary to assign class indices, which are then used to calculate accuracy. Outputs not found in the taxonomy dictionary are automatically considered incorrect. For computing frame-wise accuracy in COIN action segmentation mask, we apply a similar method.

For procedures involving multiple steps, we need to calculate step-wise accuracy. We employ a straightforward approach using string comparison to identify verb/noun indices. As noted in our training prompts, actions are separated by a semicolon ``;''. Thus, we split the model-generated content using this delimiter to extract the texts corresponding to the 20 steps.

\noindent\textbf{Ego4D LTA.} 
The Ego4D LTA benchmark, as outlined in~\cite{ego4d}, utilizes Edit Distance (ED) as its evaluation metric, as described in~\cite{edit_distance}. ED quantifies the minimum number of operations needed to transform one string into another. In contrast to previous works (e.g., ~\cite{egot2,icvae,hiervl,ego4d,antgpt,palm}) that used a classification paradigm and calculated ED based on predicted verb/noun indices, our Online-VideoLLM system, which exclusively generates text, presents challenges in metric calculation. Additionally, the method we used for evaluating on COIN Benchmarks is limited to producing results for either a single step or an overall procedure, not for more complex text outputs.

To derive verb/noun indices from our model's outputs, we use a straightforward method involving string splitting and comparison. As outlined in our training prompts, actions are separated by a semicolon ``;''. We use this delimiter to split the model-generated content into the text for each of the 20 steps. If the split results in more or fewer than 20 steps, we adjust by adding 'none' for padding or by clamping the excess steps, respectively. Next, we construct a dictionary that maps action text to their corresponding verb/noun category indices, a task facilitated by the available taxonomy annotations. Finally, this dictionary is used to convert the generated text into verb/noun category indices, which are then employed to calculate the Edit Distance (ED).

\section{More Results}\label{supp:more_results}

\noindent\textbf{Streaming Dialogue}. As shown in Table~\ref{table:coin_ego4d_stream}, we evaluate our model on joint COIN and Ego4D streaming set. COIN Stream is built by our streaming dialogue generation method, while the Ego4D narration stream simulates Ego4D annotators to write the narration while watching the video~\cite{ego4d}. From the table, we can see our method has the similar language modeling ability (reflected by LM-PPL) with the per-frame video-language dialogue format, but achieves huge advantages in fluency and time difference, which suggests better support for streaming videos.

\noindent\textbf{Demo Results with More Tokens.} Figure~\ref{fig_supp:demo} shows our demo results, supported by model trained with $1+3\times3$ tokens per frame. Though we do not show evaluation performance for more spatial tokens in our paper, we observe their quantitative results are much better than 1 token. We will update the results in our github repository. 

\begin{table}[t]
\centering
\small
\begin{tabular}{l|ccc}
\toprule
\multirow{2}{*}{Method} & \multicolumn{3}{c}{COIN + Ego4D Stream Validation} \\
&\textit{LM-PPL}$\downarrow$   & \textit{TimeDiff}$\downarrow$ &  \textit{Fluency}$\uparrow$ \\
\midrule
Per-frame Dial. & 3.29 & 6.98 & 32.9\% \\
\baseline{LIVE} & \baseline{\textbf{2.56}} & \baseline{\textbf{4.21}} & \baseline{\textbf{39.8\%}} \\
\bottomrule
\end{tabular}
\caption{Joint training of COIN Dialogue Stream and Ego4D Narration Stream. LIVE consistently performs better than per-frame dialogue method.}\label{table:coin_ego4d_stream}
\end{table}

\section{Limitations}\label{section_supp:limit}

Our primary limitation lies in the inadequacy of high-quality streaming dialogue data, which hinders its generalization capability. The dialogues generated in our method are derived from existing video datasets, which cannot capture the complex and varied requirements of real-world users. We observe the method can overfit when training on a small dataset. Our future efforts are scaling the method on larger datasets~\cite{lf-vila,vidchapters} or ASR texts in streaming video. Furthermore, we also find that the spatial ability is not strong due to its less spatial token. In the future, we will seek better trade-off strategy to balance spatial and temporal dimensions in video streaming dialogue.

\end{document}